\documentclass[10pt,twocolumn,letterpaper]{article}

\usepackage[pagenumbers]{cvpr} % To force page numbers, e.g. for an arXiv version

\usepackage{graphicx}
\usepackage{amsmath}
\usepackage{amssymb}
\usepackage{booktabs}
\usepackage{diagbox}
\usepackage{enumerate}
\usepackage{multirow}

\usepackage{cite}
\usepackage{graphicx}
\usepackage{amsmath}
\usepackage{amssymb}
\usepackage{booktabs}
\usepackage{diagbox}
\usepackage{enumerate}
\usepackage{multirow}
\usepackage{subcaption}
\usepackage{algorithm}
\usepackage{algorithmic}

\usepackage{algorithm}
\usepackage{algorithmic}
\usepackage[pagebackref,breaklinks,colorlinks]{hyperref}

\usepackage[capitalize]{cleveref}
\crefname{section}{Sec.}{Secs.}
\Crefname{section}{Section}{Sections}
\Crefname{table}{Table}{Tables}
\crefname{table}{Tab.}{Tabs.}

\begin{document}

%%%%%%%%% TITLE - PLEASE UPDATE
\title{STN: Scalable Tensorizing Networks via Structure-Aware Training and Adaptive Compression} 

\author{Chang Nie$^1$, Huan Wang$^{1*}$, Lu Zhao$^1$\\
Nanjing University of Science and Technology (NJUST), China\\
{\tt\small \{changnie, wanghuanphd\}$@$njust.edu.cn, zhaolu050826$@$gmail.com }
% For a paper whose authors are all at the same institution,
% omit the following lines up until the closing ``}''.
% Additional authors and addresses can be added with ``\and'',
% just like the second author.
% To save space, use either the email address or home page, not both
}

\maketitle

%%%%%%%%% ABSTRACT
\begin{abstract}
Deep neural networks (DNNs) have delivered a remarkable performance in many tasks of computer vision. However, over-parameter- ized representations of popular architectures dramatically increase their computational complexity and storage costs, and hinder their availability in edge devices with constrained resources. Regardless of many tensor decomposition (TD) methods that have been well-studied for compressing DNNs to learn compact representations, they suffer from non-negligible performance degradation in practice. In this paper, we propose Scalable Tensorizing Networks (STN), which dynamically and adaptively adjust the model size and decomposition structure without retraining. First, we account for compression during training by adding a low-rank regularizer to guarantee networks' desired low-rank characteristics in full tensor format. Then, considering network layers exhibit various low-rank structures, STN is obtained by a data-driven adaptive TD approach, for which the topological structure of decomposition per layer is learned from the pre-trained model, and the ranks are selected appropriately under specified storage constraints. As a result, STN is compatible with arbitrary network architectures and achieves higher compression performance and flexibility over other tensorizing versions. Comprehensive experiments on several popular architectures and benchmarks substantiate the superiority of our model towards improving parameter efficiency. 

\end{abstract}

%%%%%%%%% BODY TEXT

\section{Introduction}
\label{sec:intro}
Driven by new computational hardware and large-scale training data, deep neural networks (DNNs) have gained remarkable achievements in a variety of applications, including image and video recognition~\cite{ref0,ref3}, object detection~\cite{ref1,ref2}, instance segmentation~\cite{ref4,ref6}, and image generation~\cite{ref7}, to name a few. At the same time, recent studies reveal that over-parameterization is a crucial feature for successfully training DNNs, which encourages them to find better local minima in huge parameter space~\cite{ref8,ref9}. However, the over-parameterized representations of neural networks rely on excessive computational and storage costs, thereby severely restricting their usability in resource-constrained edge devices, such as mobile phones and Internet-of-Things (IoT) devices~\cite{ref10}.

Consequently, various compression techniques have been proposed for reducing the computational and memory consumption of DNNs, including low-rank approximation~\cite{ref11,ref12}, quantization~\cite{ref14,ref15}, structural sparsification~\cite{ref13}, and pruning~\cite{ref16}. The goal of these methods is to convert a high-precision large model into a small model with low complexity while maintaining or even improving the model's performance as possible. Among these strategies, the low-rank approximation based on tensor decomposition (TD) has a solid theoretical rationale~\cite{ref17,ref18} for capturing the intrinsic structure inside the weight tensors by representing them as multilinear operations over a sequence of latent factors. Moreover, TD-based approaches can achieve higher compression ratios and even obtain accuracy gains by mitigating the overfitting of a pre-trained model~\cite{ref19,ref20}.

In this light, many TD models have been exploited to eliminate redundancy and promote parameter efficiency of convolutional neural networks (CNNs), which mainly focus on reparameterizing the convolutional and fully-connected layers~\cite{ref12,ref21,ref22}. As two of the most classical TD methods, CANDECOMP/PAR- AFAC (CP) and Tucker decomposition~\cite{ref23,ref24} were used earlier to accelerate and compress CNNs~\cite{ref22,ref28}. By virtue of tensor networks (TNs)~\cite{reftn}, the recently proposed tensor train (TT)~\cite{ref25} and tensor ring (TR)~\cite{ref26} decompositions exhibit better behaviour on higher-order tensor representations. In particular, the storage complexity of TT and TR is linear with the tensor order. However, as pointed out in~\cite{ref27}, changing the model size often leads to non-negligible performance degradation, even if the low-rank approximated model is retrained over a long period. This necessitates a new compression paradigm to solve this difficulty, that is, to construct a more compact and flexible representation with little reducing the expressiveness of the original layers. To achieve that, we rethink the inherent limitations of existing TD-based compression methods. 
%inherent

(\romannumeral1) \textbf{Scalability.}
A pre-trained model requires compression into multiple versions of different sizes to match specific scenarios' computational and diverse storage requirements. However, retraining or fine-tuning each smaller model separately requires additional resources. It is desirable to resize the uncompressed model without retraining while maintaining high accuracy dynamically since decomposition computation in the post-processing stage is trivial. Thereby, several arts~\cite{ref20,ref27,ref37} try to account for compression during the training procedure, such as adding regularizers and compression loss to preserve the desired low-rank characteristic for the uncompressed models. But they either have unsatisfied performance~\cite{ref37} or require predetermining the TD model and ranks~\cite{ref20,ref27}.

(\romannumeral2) \textbf{Model Selection and Rank Determination.}
A majority of precedent studies~\cite{ref12,ref22,ref31,ref32} prefer to select a suitable decomposition model to compress the weight tensors in DNNs, e.g., the convolution kernels and dense matrix, expecting to achieve higher compression ratios while maintaining the model performance. However, varying sizes and depths of the parameters exhibit different low-rank structural characteristics, making it inappropriate to decompose all layers with a single TD model. For instance, the Tucker decomposition applies to convolutional layers well rather than fully-connected layers, while the TT is the opposite~\cite{ref30,ref32}. In addition, it is laborious and cumbersome to determine the TD-ranks for each layer to satisfy the memory requirements. Hence, the suboptimal model selection and hand-designed ranks lead to non-negligible performance degradation.

\begin{figure*}[ht]
  \centering
   \includegraphics[width=0.85\linewidth]{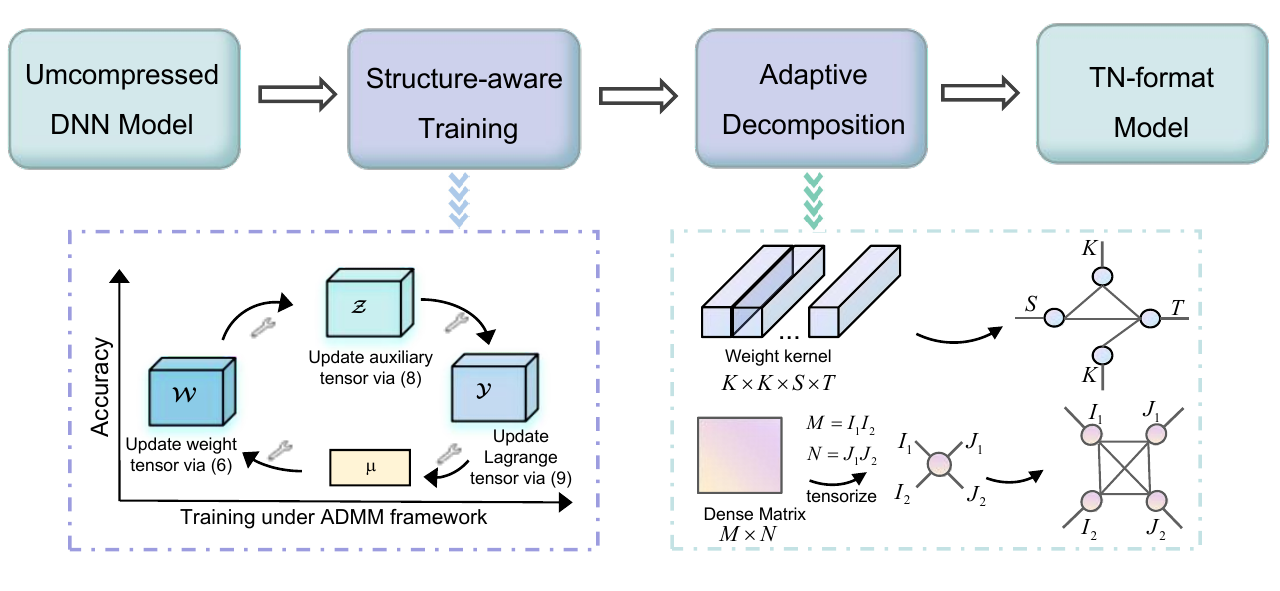}
   \vspace{-5mm}
   \caption{Overall procedure of the Scalable Tensorizing Networks (STN). Note that the training and decomposition schemes are suitable for all kinds of network architectures.\vspace{-5mm}}
   \label{img2}
\end{figure*}

To tackle the existing limitations listed above, we propose a novel scalable compression framework to eliminate the redundancy of neural networks (shown in Fig.~\ref{img2}). First, we present a structure-aware training scheme that introduces an additional low-rank regularizer during the training procedure for the uncompressed model to ensure the desired low-rank characteristics for reducing the approximation-caused error. That is separate from~\cite{ref20,ref27} because we are unnecessary to specify the TD model and rank in advance. To further compress the pre-trained model, a data-driven adaptive TD strategy is introduced to simultaneously compress the convolutional and fully-connected layers of DNNs, which enables us to learn the most appropriate decomposition topology for each weight tensor and to select the proper ranks under specified storage constraints (see Fig.~\ref{img5}). Note that the adaptive decomposition strategy favours the kernels with unbalanced mode dimensions, and this is the main difference from other TD-based approaches~\cite{ref12,ref22,ref30,ref31} with fixed topologies and manually designed ranks. Conclusively, the main technical contributions of our work can be summarized as follows:%without retraining

\begin{itemize}
\item We propose STN to eliminate redundancy of overparameterized DNNs and improve their scalability, allowing flexibility to change the model size without retraining.

\item The training of STN is performed under alternating direction method of multipliers (ADMM)~\cite{ref40} framework by solving a regularized optimization problem. That maintains the performance of the full-rank model, and the obtained inherent low-rank characteristic reduces the approximation error.

\item An alternating least squares (ALS)-based scalable compression algorithm is presented to tensorize STN, which avoids the problems of model selection and rank determination while capturing the intrinsic global structure of the weight tensors better.

\item We conduct extensive experiments on several prominent architectures and benchmarks to show the superiority of STN over other scalable and tensorizing counterparts. 
% In particular, the ADN achieves 3.5$\times$ compression and 4.5$\times$ FLOPs reduction for compressing ResNet-56 on CIFAR10 and ResNet-18 on ImageNet, respectively, with little degradation in accuracy, and RSN yields competitive accuracy over a wide range of model sizes.

\end{itemize}

\noindent\textbf{Notations.} In this paper, the tensors, matrices, vectors, and scalars are denoted as $\boldsymbol{\mathcal{A}}, \boldsymbol{{A}}, \boldsymbol{{a}}$, and $a$, respectively. The $[\boldsymbol{k}]$ represents a set including integers from 1 to $k$. For an $d$-order tensor $\boldsymbol{\mathcal{A}}\in \mathbb{R}^{I_1\times\cdots\times I_d}$, its $(i_1,\cdots,i_d)$th entry is denoted as $\boldsymbol{\mathcal{A}}(i_1,\cdots,i_d)$ or $\boldsymbol{\mathcal{A}}_{i_1,\cdots,i_d}$. The inner product of tensors $\boldsymbol{\mathcal{A}},\boldsymbol{\mathcal{B}}\in \mathbb{R}^{I_1\times\cdots\times I_d}$ is defined as $\left \langle \boldsymbol{\mathcal{A}},\boldsymbol{\mathcal{B}} \right \rangle = \sum_{i_1,\cdots,i_d}^{I_1,\cdots,I_d} \boldsymbol{\mathcal{A}}_{i_1,\cdots,i_d}\boldsymbol{\mathcal{B}}_{i_1,\cdots,i_d}$, and $||\boldsymbol{\mathcal{A}}||_F=\sqrt{\left \langle \boldsymbol{\mathcal{A}},\boldsymbol{\mathcal{A}} \right \rangle}$ denotes Frobenius norm of $\boldsymbol{\mathcal{A}}$. Plus, the $k$-unfolding of $\boldsymbol{\mathcal{A}}$ is denoted as $\boldsymbol{A}_{(k)}\in \mathbb{R}^{I_k\times \prod_{k'\neq k}I_{k'}}$, which is a matrix obtained by concatenating all the mode-$k$ fibers. For a matrix $\boldsymbol{\mathcal{A}}\in \mathbb{R}^{I_1\times I_2}$, its nuclear norm $||\boldsymbol{\mathcal{A}}||_*$ is defined as the sum of singular values of $\boldsymbol{\mathcal{A}}$. More notations see literature~\cite{ref17}.

\section{Related Work}
\noindent\textbf{Scalable networks}~\cite{ref38,ref46,ref47,ref27} aim to directly compress a well-trained DNN model to achieve high performance without retraining. However, the weight tensors of an over-parameterized representation may not exhibit apparent structural low-rankness in general, and directly decomposing a full-rank model into the a low-rank format would inevitably introduce significant approximation errors. Indeed, the phenomenon of layer-wise approximation inconsistency is accumulated as the depth of the uncompressed model increases. In~\cite{ref38,ref46}, the authors implicitly account for compression in training by adding a low-rank regularizer to ensure the desired properties of the model. Yu et al.~\cite{ref47} proposed a slimmable network that can dynamically adjust its width to meet practical needs. Moreover, Yaguchi et al.~\cite{ref27} introduced a Decomposable-Net jointly minimizing both full-rank and low-rank network losses and exploiting the fact that the nuclear norm of a matrix can be matched by the Frobenius norm of two factorized factors.

\vspace{2mm}
\noindent\textbf{Low-rank approximation} methods are broadly investigated to compress DNNs by their solid theoretical rationale and effectiveness. Typically, TD-based compressions~\cite{ref12,ref22,ref28} represent the large-size weight tensor into a multilinear operation within a sequence of latent factors and then reparameterize the existing layers, such as convolutional and fully-connected layers, to accelerate and compress the DNN model. Several earlier works~\cite{ref11,ref41} proposed using truncated singular value decomposition (SVD) to compress the dense matrix in classification layers. Later, \cite{ref22} and \cite{ref28} suggest using CP and Tucker decomposition to compress the convolution kernel by replacing the original layers with several consecutive ones having smaller kernels and then fine-tuning the resulting network to recover its accuracy. Besides, the advanced TT~\cite{ref25} and TR decompositions~\cite{ref26} introduced by the aid of tensor networks are more suitable for higher-order tensor representations, and tensorizing the primal low-order weights before decomposing enable to achieve higher compression ratios. Notably, Wang et al.~\cite{ref31} used TR to improve the parameter efficiency of ResNet-32 on CIFAR10 and obtained $5\times$ speedup with the expense of 1.9\% accuracy drop. However, existing methods compress each layer uniformly with the same scheme while ignoring their different structural characteristics and importance, which leads to significant performance degradation in practice.

% \noindent\textbf{Other methods for DNN compression.} Quantization methods~\cite{ref14, ref15} offer a lightweight representation of the uncompressed model by directly expressing parameters or activations with discrete lower-bits values. As an example, \cite{ref14} used 8-bits and 5-bits to quantize 32-bits of parameters in the convolutional and fully-connected layers, respectively. Notably, the binary quantization strategy~\cite{ref34} restricts the weights to values of 1 or -1, and then accelerates the neural network by hardware-friendly binary operations. However, the choice of quantization interval affects the speedup effect in practice, and the resulting compression ratio is limited (up to $32\times$)~\cite{ref20}. Moreover, pruning strategies are designed to speed up the network by removing redundant components of DNNs, including weight pruning~\cite{ref16}, channel pruning~\cite{ref35}, and layer pruning~\cite{ref36}. The unstructured weight pruning cuts out unimportant connections with small weight values between the neurons. Also, introducing a sparse regularizer (e.g., $\ell_1$ loss) in training is beneficial to maintain the performance after pruning~\cite{ref42}. However, the unstructured pruning reduces the expected speedup and is not suitable for parallel processing. In contrast, the structured channel and layer pruning are preferred in practice but suffer from significant performance decrease.

\vspace{-4mm}
\section{Scalable Tensorizing Networks}
This section details the proposed STN for improving the scalability and compactness of neural networks. First, we offer a novel training scheme for STN learning under the ADMM framework~\cite{ref40}. Then, to scale the pre-trained low-rank model in full-tensor format, we introduce a generalized and adaptive TD method and apply it to simultaneously reparameterize the convolutional and fully-connected layers.

\vspace{-4mm}
\subsection{Structure-aware Training Scheme}
In general, a pre-trained uncompressed DNN model rarely exhibits obvious low-rankness, and decomposing it to TN-format (e.g., Tucker, TT, and TR) would inevitably cause significant approximation errors~\cite{ref20}. To improve the scalability of neural networks, we introduce a structure-aware training solution (see Fig.~\ref{img2}) to guarantee the desired low-rank characteristics and thus avoid the subsequent retraining phase.

\vspace{2mm}
\noindent\textbf{Problem Formulation.}
For a network with $L$ layers, let $\mathcal{W}=\{\mathcal{W}_l\}_{l=1}^L$ be the set of weight tensors. The network learning can be formulated as a joint minimization optimization problem of the objective function and the regularization item as follows
\begin{equation}\begin{split}
\min_{\boldsymbol{\mathcal{W}}} \ell(\boldsymbol{\mathcal{W}}) +\lambda \sum_{l=1}^L\Omega(\boldsymbol{\mathcal{W}}_l),
\label{eq7}
\end{split}\end{equation}
where $\ell(\cdot)$ is a loss function, e.g.,  the cross-entropy, and $\Omega(\cdot):\mathbb{R}^{I_1\times\cdots\times I_d}\to \mathbb{R}^+$ is a regularizer that ensures the low-rankness of $\boldsymbol{\mathcal{W}}$, and $\lambda\in \mathbb{R}^+$ is a trade-off constant to balance the two terms.

\vspace{2mm}
\noindent\textbf{Optimization Framework.}
Due to the optimization of tensor rank being NP-hard, many approaches extend the tensor low-rank to the unfolding matrix setting and then solve it using off-the-shelf algorithms. Examples include mode unfolding~\cite{ref17} and generalized unfolding~\cite{ref43}. Then, as the tightest convex relaxation of the matrix rank, the nuclear norm is frequently suggested for $\Omega(\boldsymbol{\mathcal{W}})$. The alternative of (\ref{eq7}) can be written as
\begin{equation}\begin{split}
\min_{\boldsymbol{\mathcal{W}}} \ell(\boldsymbol{\mathcal{W}}) +\lambda \sum_{l=1}^L ||\zeta (\boldsymbol{\mathcal{W}}_{l})||_*,
\label{eq8}
\end{split}\end{equation}
where $\zeta (\boldsymbol{\mathcal{W}}_l)$ is the unfolding operation~\cite{ref44} to obtain a balanced matrix of $\boldsymbol{\mathcal{W}}_l$. Then, we introduce auxiliary variable, $\boldsymbol{\mathcal{Z}}=\{\boldsymbol{\mathcal{Z}}_l\})_{l=1}^L$, and rewrite (\ref{eq8}) as:
\begin{equation}\begin{split}
&\min_{\boldsymbol{\mathcal{W}}, \boldsymbol{\mathcal{Z}}} \ell(\boldsymbol{\mathcal{W}}) +\lambda \sum_{l=1}^L ||\zeta (\boldsymbol{\mathcal{Z}}_{l})||_*\\
& s.t. \ \ \boldsymbol{\mathcal{Z}}_l=\boldsymbol{\mathcal{W}}_l , l=1,\cdots,L.
\label{eq9}
\end{split}\end{equation}
By integrating the equal constraints into the objective optimization function, the augmented Lagrangian version of (\ref{eq9}) can be derived as\vspace{-2mm}
\begin{equation}\begin{split}
\mathcal{L}(\boldsymbol{\mathcal{W}}, \boldsymbol{\mathcal{Z}},\boldsymbol{\mathcal{Y}})= \ell(\boldsymbol{\mathcal{W}}) + & \lambda \sum_{l=1}^L \biggl\{  ||\zeta (\boldsymbol{\mathcal{Z}}_{l})||_* \\
+ \left \langle \boldsymbol{\mathcal{Y}}_l, \boldsymbol{\mathcal{Z}}_l - \boldsymbol{\mathcal{W}}_l \right \rangle
&+ \frac{\mu}{2}||\boldsymbol{\mathcal{Z}}_l- \boldsymbol{\mathcal{W}}_l ||_F^2  \biggl\},
\label{eq10}
\end{split}\end{equation}
where $\boldsymbol{\mathcal{Y}}=\{\boldsymbol{\mathcal{Y}}_l\})_{l=1}^L$ is Lagrangian multiplier tensor set and $\mu$ is a tuning parameter. The above model is non-convex differentiable and the variables $\boldsymbol{\mathcal{W}},\boldsymbol{\mathcal{Z}}$ and $\boldsymbol{\mathcal{Y}}$ are independent. The solution of (\ref{eq10}) can be obtained by iteratively updating each variable while fixing the others as the following mechanism.

\vspace{2mm}
1) Update $\boldsymbol{\mathcal{W}}_l$:  Retrieve all terms in (\ref{eq10}) associated with $\boldsymbol{\mathcal{W}}_l$, and then the $\boldsymbol{\mathcal{W}}_l$ $(l=1,\cdots,L)$-subproblem can be formulated as
\begin{equation}\begin{split}
\min_{\boldsymbol{\mathcal{W}}_l}\ \ \ell(\boldsymbol{\mathcal{W}}) +  \frac{\lambda\mu}{2}||\boldsymbol{\mathcal{Z}}_l-\boldsymbol{\mathcal{W}}_l + \frac{1}{\mu} \boldsymbol{\mathcal{Y}}_l ||_F^2.
\label{eq11}
\end{split}\end{equation}
In fact, the loss function $\ell(\cdot)$ is routinely optimized with standard stochastic gradient descent (SGD) algorithm. Then $\boldsymbol{\mathcal{W}_l}$ can be updated by
\begin{equation}\begin{split}
\boldsymbol{\mathcal{W}}_l^{(s+1)}=& \boldsymbol{\mathcal{W}}_l^{(s)} - \eta \frac{\partial \mathcal{L}(\boldsymbol{\mathcal{W}}, \boldsymbol{\mathcal{Z}}^{(s)}_l,\boldsymbol{\mathcal{Y}}^{(s)}_l)}{\partial \boldsymbol{\mathcal{W}}^{(s)}_l}= \boldsymbol{\mathcal{W}}_l^{(s)} -\\
& \eta(\frac{\partial \ell(\boldsymbol{\mathcal{W}})}{\partial \boldsymbol{\mathcal{W}}_l^{(s)} }  + \lambda\mu (\boldsymbol{\mathcal{W}}_l^{(s)} - \boldsymbol{\mathcal{Z}}^{(s)}_l- \frac{1}{\mu} \boldsymbol{\mathcal{Y}}^{(s)}_l),
\label{eq12}
\end{split}\end{equation}
here $s$ and $\eta$ denote the iterative step and learning rate, respectively.

\vspace{2mm}
2) Update $\boldsymbol{\mathcal{Z}}_l$: The $\boldsymbol{\mathcal{Z}}_l$ $(l=1,\cdots,L)$-subproblem can be expressed as
\begin{equation}\begin{split}
\min_{\boldsymbol{\mathcal{Z}}_l}\ \ ||\zeta (\mathcal{Z}_{l})||_* + \frac{\mu}{2}||\boldsymbol{\mathcal{Z}}_l- \boldsymbol{\mathcal{W}}_l + \frac{1}{\mu}\boldsymbol{\mathcal{Y}}_l ||_F^2.
\label{eq13}
\end{split}\end{equation}
Let $\mathcal{D}_{1/\mu}(\cdot)$ be the singular value thresholding operation~\cite{ref45}, then the closed-form solution of the above formulation can be computed by
\begin{equation}\begin{split}
\boldsymbol{\mathcal{Z}}_l^{(s+1)} = \zeta^{-1}\biggl\{\mathcal{D}_{1/\mu} \Bigl[ \zeta(\boldsymbol{\mathcal{W}}_l^{(s)} - \frac{1}{\mu}\boldsymbol{\mathcal{Y}}_l^{(s)}) \Bigl] \biggl\}.
\label{eq14}
\end{split}\end{equation}

3) Update $\boldsymbol{\mathcal{Y}}_l$: The update rule for Lagrange multipliers $\boldsymbol{\mathcal{Y}}_l$ $(l=1,\cdots,L)$ is given by
\begin{equation}\begin{split}
\boldsymbol{\mathcal{Y}}_l^{(s+1)} = \boldsymbol{\mathcal{Y}}_l^{(s+1)} + \mu (\boldsymbol{\mathcal{Z}}_l^{(s)} - \boldsymbol{\mathcal{W}}_l^{(s)}).
\label{eq15}
\end{split}\end{equation}
In particular, the parameter $\mu$ is updated at each iteration via $\mu=\min\{\rho\mu, \mu_{max}\}$, where $\rho$ is a tuning hyperparameter. Since the number of iterations of the ADMM algorithm is usually much less than $\ell(\cdot)$, the SVD operation involved in (\ref{eq14}) is computationally expensive. We adopt a periodic fashion to train the structure-aware DNN model, that is, each ADMM-iteration is performed once in every $m$ regular SGD steps. The overall process of training STN is summarized in Algorithm~\ref{alg2}, and its convergence analysis are empirically verified in the experimental section.

\floatname{algorithm}{Algorithm}
\renewcommand{\algorithmicrequire}{\textbf{Input:}}  
\renewcommand{\algorithmicensure}{\textbf{Output:}}    
	\begin{algorithm}[!t]
		\caption{: Training procedure of STN}
		\begin{algorithmic}[1]
			\label{alg2}
			\REQUIRE Weight tensors $\boldsymbol{\mathcal{W}}$,  trade-off constant $\lambda $, maximum iterations $S_{\max}$, period $m=100$.\\
			\ENSURE Optimized $\boldsymbol{\mathcal{W}}$.
			\STATE Initialize $\boldsymbol{\mathcal{W}}, \mu=1, \rho=1.001, \mu_{\max}=10$.\\
			\STATE $\boldsymbol{\mathcal{Z}}\gets\boldsymbol{\mathcal{W}}$, $\boldsymbol{\mathcal{Y}}\gets 0$.
			\WHILE{$s\leq S_{\max}$}
			\IF{$s\ \%\ m\neq0$}
			\STATE Update $\boldsymbol{\mathcal{W}}$ via SGD.
			\ELSE
			\STATE Update $\boldsymbol{\mathcal{W}}$ via (\ref{eq12}).
			\STATE Update $\boldsymbol{\mathcal{Z}}$ via (\ref{eq14}).
			\STATE Update $\boldsymbol{\mathcal{Y}}$ via (\ref{eq15}).
			\STATE $\mu\gets \min(\rho\mu, \mu_{\max})$.
			\ENDIF
			\ENDWHILE
		\end{algorithmic}
\end{algorithm}

\subsection{Tensor Network Decomposition}

Before compressing the pre-trained model obtained via Algorithm~\ref{alg2}, we necessitate introducing a generalized TN decomposition model~\cite{ref39} to fully leverage the low-rank prior. For an $N$-order tensor $\boldsymbol{\mathcal{X}}$ of size $I_1\times \cdots\times I_N$, the TN decomposition represents it as a multilinear operation over $N$ latent factors as $\boldsymbol{\mathcal{X}}=\Re(\boldsymbol{\mathcal{Z}}^{(1)}, \cdots, \boldsymbol{\mathcal{Z}}^{(N)})$, where $\Re$ is the tensor contraction operator~\cite{ref48} and $\boldsymbol{\mathcal{Z}}^{(k )}\in \mathbb{R}^{R_{1,k}\times \dots \times R_{k-1, k}\times I_k \times R_{k+1, k}\times \dots\times R_{ N,k}}$ with $k\in [N]$. Specifically, the element-wise form can be formulated as
\begin{equation}\begin{split}
\resizebox{0.9\hsize}{!}{$\begin{aligned}
&\boldsymbol{\mathcal{X}}(i_1,i_2,...,i_N)=\sum_{r_{1,2}=1}^{R_{1,2}}\cdots\sum_{r_{1,N}=1}^{R_{1,N}}\sum_{r_{2,3}=1}^{R_{2,3}}\cdots\sum_{r_{2,N}=1}^{R_{2,N}}\cdots\sum_{r_{N-1,N}=1}^{R_{N-1,N}}\\
&\Bigl\{\boldsymbol{\mathcal{Z}}^{(1)}({i_1,r_{1,2},...,r_{1,N}})\boldsymbol{\mathcal{Z}}^{(2)}({r_{1,2},i_2,...,r_{2,N}})\cdots\boldsymbol{\mathcal{Z}}^{(N)}({r_{1,N},...,r_{N-1,N},i_N})\Bigl\}.
\label{eq1}
\end{aligned}$}
\end{split}\end{equation}
The vector $[R_{1,2},\cdots,R_{1,N}, R_{2,3},\cdots,R_{2,N},\cdots,R_{N-1,N}]\in \mathbb{N}_+^{\frac{N(N-1)}{2}}$ is called TN-ranks. 
%Essentially, a TN model can be considered as a fully-connected network topology (see Fig.~\ref{img1}), which constructs a connection with rank $R_{m,n}$ for any two factors $\boldsymbol{\mathcal{Z}}^{(m)}$ and $\boldsymbol{\mathcal{Z}}^{(n)}$ and contracts all standard edges to obtain the target tensor~\cite{ref48}. Note that the existence of rank-1 connections is non-influential for keeping computation consistency, 
Essentially, rank-1 connections are non-influential for keeping computation consistency, and many typical TD models can be considered as generalizations of the TN decomposition, including TT~\cite{ref25} and TR~\cite{ref26}. Hence, the TN-ranks determine the storage complexity and topological structure of the TN-format tensor.

% \begin{figure}[t]
%   \centering
%   \includegraphics[width=0.6\linewidth]{fig/1.pdf}
%   \caption{A graphical representation of the tensor network decomposition.\vspace{-5mm}}
%   \label{img1}
% \end{figure}

\vspace{2mm}
\noindent\textbf{Rank Determination.}
In the decomposition stage of DNNs, selecting suitable TN-ranks for each weight tensor is critical to maximizing performance. The previous TT- and TR-based methods layer-wisely determine the rank manually. However, this is time-consuming and laborious for neural networks containing tens or even hundreds of layers. Realizing the fact that mode correlations of the approximated tensor $\widetilde{\boldsymbol{\mathcal{X}}}\in\mathbb{R}^{I_1\times \cdots\times I_N}$ are established based on corresponding factor connections, we leverage the below formulations from~\cite{ref39} to determine the TN-ranks
\begin{equation}\begin{split}
% \overset{\text{def}}{=}
R_{m,n}=\min_{x}\ \  \frac{||\sum_{k=1}^{C} (\pmb{S}_k^{(m, n)})_{1:x}||_F^2}{||\sum_{k=1}^{C} \pmb{S}_k^{(m, n)}||_F^2} \geq\kappa,
\label{eq2}
\end{split}\end{equation}
where $1\leq m<n\leq N$, $C=\prod_{i\neq m,n} I_i$, $\pmb{S}_k^{(m, n)}$ is a vector comprising the singular values of the $k$th frontal slice of $\boldsymbol{\mathcal{X}}_{(m,n)}\in\mathbb{R}^{I_m\times I_n\times  \prod_{i\neq m,n} I_i}$, and the constant $\kappa$ $(0<\kappa\leq 1)$ can be viewed as the degree of information retention among modes correlations. Then the TN-rank can be determined adaptively according to the mode correlations of the instance tensor beyond the manually setting.

\begin{figure*}[!t]
  \centering
   \includegraphics[width=0.85\linewidth]{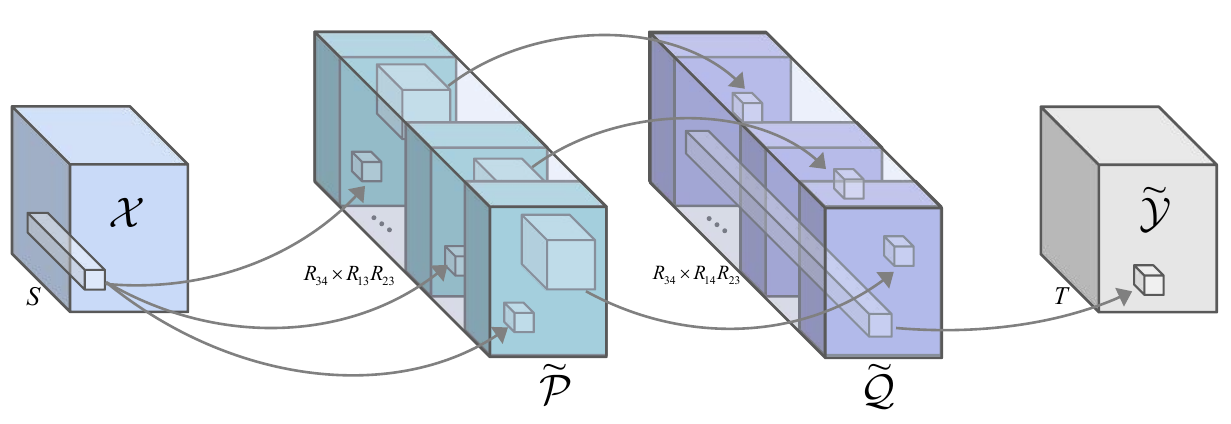}
    \vspace{-2mm}
   \caption{Flowchart of TN-format convolution layer. Note that the kernels for the parameter-shared group convolution of the intermediate process are generated with factor contraction (Eq.(\ref{eq5})).}
   \label{img_conv}
\end{figure*}

\subsection{Tensorizing Neural Networks}
%In this part, we directly decompose a large uncompressed low-rank model into smaller versions in TN-format with high-precision. In general, weight tensors of different layers in DNNs exhibit varied low-rank behaviour and suffer from mode dimensional unbalance problems, e.g., the dimension of channel mode of the kernels is much larger than the spatial ones. As a result, the existing low-rank compression methods using a single TD model to handle all network layers are inappropriate. To overcome the hardness of model selection, we apply the aforementioned TD strategy to perform low-rank approximation after obtaining an uncompressed DNN model in full-tensor format.
\noindent\textbf{Convolutional Layer Compression.}
The convolutional layer is the standard operation unit of modern CNNs. Given a 3-order input tensor $\boldsymbol{\mathcal{X}}\in\mathbb{R}^{W\times H\times S}$, which is convoluted by a 4-order weight tensor $\boldsymbol{\mathcal{K}}\in\mathbb{R}^{K\times K\times S\times T}$ to yield the 3-order output tensor $\boldsymbol{\mathcal{Y}}\in\mathbb{R}^{(W-K+1)\times (H-K+1)\times T}$ with the following linear mapping:
\begin{equation}\begin{split}
% \left \langle \boldsymbol{\mathcal{X}}_{(w:w+K, h:h+K)}, \boldsymbol{\mathcal{K}}_{(:,:,:,t)} \right \rangle
\boldsymbol{\mathcal{Y}}(w, h, t)=  \sum_{k_1=1}^K \sum_{k_2=1}^K \sum_{s=1}^S\Bigl\{&\boldsymbol{\mathcal{X}}(k_1+w-1,k_2+h-1, s)\\
&\boldsymbol{\mathcal{K}}(k_1,k_2,s,t)\Bigl\}.
\label{eq4}
\end{split}\end{equation}
Suppose that $\boldsymbol{\mathcal{K}}$ can be approximated by $\widetilde{\boldsymbol{\mathcal{K}}}$ in TN-format as (\ref{eq1}), $\widetilde{\boldsymbol{\mathcal{K}}}=\Re(\boldsymbol{\mathcal{Z}}^{(1)},\cdots, \boldsymbol{\mathcal{Z}}^{(4)})$ with proper TN-ranks $(R_{1,2},R_{1,3}, R_{1,4},R_{2,3},R_{2,4},R_{3,4})$. Then the new output tensor $\widetilde{\boldsymbol{\mathcal{Y}}}$ can be simply computed by
\begin{equation}
\begin{split}
&\widetilde{\boldsymbol{\mathcal{P}}}_{w', h', r_{1,3}, r_{2,3},r_{3,4}}=\sum_{s=1}^S \boldsymbol{\mathcal{X}}_{w',h',s}\boldsymbol{\mathcal{Z}}^{(3)}_{r_{1,3},r_{2,3}, s, r_{3,4}},\\
&\widetilde{\boldsymbol{\mathcal{Q}}}_{w, h, r_{1,4}, r_{2,4},r_{3,4}}=  \sum_{k_1=1}^K \sum_{k_2=1}^K \sum_{r_{1,3}}^{R_{1,3}} \sum_{r_{2,3}}^{R_{2,3}}\bigl(\sum_{r_{1,2}}^{R_{1,2}} \boldsymbol{\mathcal{Z}}^{(1)}_{k_1, r_{1,2}, r_{1,3}, r_{1,4}}\\
&\qquad\qquad\boldsymbol{\mathcal{Z}}^{(2)}_{r_{1,2}, k_2, r_{2,3}, r_{2,4}}\bigl)\widetilde{\boldsymbol{\mathcal{P}}}_{k_1+w-1,k_2+h-1,
r_{1,3}, r_{2,3}, r_{3,4}}\\
&\widetilde{\boldsymbol{\mathcal{Y}}}(w, h, t)=\sum_{r_{1,4}}^{R_{1,4}}\sum_{r_{2,4}}^{R_{2,4}}\sum_{r_{3,4}}^{R_{3,4}}\widetilde{\boldsymbol{\mathcal{Q}}}_{w, h, r_{1,4}, r_{2,4},r_{3,4}}\boldsymbol{\mathcal{Z}}^{(4)}_{r_{1,4},r_{2,4},r_{3,4},t}.\\
\label{eq5}
\end{split}
\end{equation}

\vspace{-5mm}
\noindent Since the merging order affects the computational complexity of contraction~\cite{ref31}, the factors are sequentially acting on the input tensor $\boldsymbol{\mathcal{X}}$ follow the order of (\ref{eq5}) in the implementation process. Then the original convolutional layer can be replaced by a few new ones with smaller kernels, whereas the reduction ratio in parameter ($P_{conv}$) and FLOPs ($C_{conv}$) are
\begin{equation}
\begin{split}
\resizebox{0.9\hsize}{!}{$\begin{aligned}
&P_{conv}=\frac{K^2ST}{(2K+S+T)R^3}\\
&C_{conv}=\frac{K^2STWH}{ WH(S+T)R^3 + WHK^2R^5 + K^2R^5 }.
\label{conv_conpl}
\end{aligned}$}
\end{split}
\end{equation}

%In addition, we can insert nonlinear activations into the tensorized module to improve its expressiveness~\cite{ref33}, since the descending and ascending of channel dimension implicitly degrades the rank of the feature maps

\floatname{algorithm}{Algorithm}
\renewcommand{\algorithmicrequire}{\textbf{Input:}}  
\renewcommand{\algorithmicensure}{\textbf{Output:}}    
	\begin{algorithm}[!t]
		\caption{: ALS-Based Compression Algorithm}
		\begin{algorithmic}[1]
			\label{alg1}
			\REQUIRE Pre-trained weights $\boldsymbol{\mathcal{W}}=\{\boldsymbol{\mathcal{W}}_l\})_{l=1}^L$, compression ratios $r$.\\
			\ENSURE Factors $\{\boldsymbol{\mathcal{Z}}_l^{(1)},\cdots,\boldsymbol{\mathcal{Z}}_l^{(N)}\}_{l=1}^L$.
			\STATE Calculate $\kappa$ and TN-ranks $R^*$.
			\STATE Initialize $\boldsymbol{\mathcal{Z}}^{(n)}_l$, relative error $\epsilon$=$1e-5$, $S_{\max}=300$.\\
			\FOR{$l$=1 to $L$}
			\STATE $s\gets 1$.
			\REPEAT
			\STATE $\widetilde{\boldsymbol{\mathcal{W}}_l}\gets\Re(\boldsymbol{\mathcal{Z}}^{(1)}_l,\cdots,\boldsymbol{\mathcal{Z}}^{(N)}_l)$.
			\FOR{$n$=1 to $N$}
			\STATE ${\boldsymbol{{Z}}}_l^{(\neq n)}\gets \texttt{reshape}(\Re(\{\boldsymbol{\mathcal{Z}}^{(i)}_l\}_{i\neq n}), \bigl[\prod_{i\neq n}I_i, -1\bigl])$.\\
			\STATE ${\boldsymbol{{Z}}}^{(n)}_l\gets  \boldsymbol{X}_{(n)}  \boldsymbol{Z}_l^{(\neq n)} (\boldsymbol{Z}_l^{(\neq n)T}{\boldsymbol{{Z}}}_l^{(\neq n)})^{\dagger}$.
			\STATE ${\boldsymbol{\mathcal{Z}}}^{(n)}_l \gets \texttt{permu}(\texttt{reshape}(\boldsymbol{{Z}}^{( n)}_l, \bigl[ I_n,R_{1,n},$\\	$\cdots,R_{n-1,n}, R_{n,n+1}, \cdots,R_{n,N}   \bigl]))$.\\
			\ENDFOR
			\STATE $rse\gets \frac{||\Re(\boldsymbol{\mathcal{Z}}^{(1)}_l,\cdots, \boldsymbol{\mathcal{Z}}^{(N)}_l)  - \widetilde{\boldsymbol{\mathcal{W}}}_l ||_F}{||\widetilde{\boldsymbol{\mathcal{W}}}_l||_F}$
			\STATE $ s\gets s+1$.
			\UNTIL $s\geq S_{\max}$ or convergence condition $rse \leq \epsilon$ is achieved.
			\ENDFOR
		\end{algorithmic}
\end{algorithm}

\vspace{2mm}
\noindent\textbf{Fully-Connected Layer Compression.}
The fully-connected layer contains a linear transformation defined by a large dense matrix $\boldsymbol{W}\in\mathbb{R}^{M\times N}$. Given an input high-dimensional signal $\boldsymbol{x}\in\mathbb{R}^{N}$, the output signal $\boldsymbol{y}\in\mathbb{R}^{M}$ can be computed by $\boldsymbol{y}=\boldsymbol{W}\boldsymbol{x}$. We first tensorize $\boldsymbol{W}$ to a high-order tensor $\boldsymbol{\mathcal{W}}\in\mathbb{R}^{I_1\times\cdots\times I_m\times J_1\times\cdots\times J_n}$, where $M=\prod_{i=1}^m I_i, N=\prod_{j=1}^n J_i$. Then, $\boldsymbol{\mathcal{W}}$ is represented in a compact multilinear TN-format by (\ref{eq1}), and we have
\begin{equation}\begin{split}
&\boldsymbol{y}(\overline{i_1\cdots i_m})= \sum_{j_1,\cdots,j_n}^{J_1,\cdots,J_n} \boldsymbol{x}(\overline{j_1\cdots j_n})\Re(\boldsymbol{\mathcal{Z}}^{(1)}_{(i_1:\cdots)},\\
&\cdots,\boldsymbol{\mathcal{Z}}^{(m)}_{(\cdots:i_m:\cdots)},  \boldsymbol{\mathcal{Z}}^{(m+1)}_{(\cdots : j_1 : \cdots)},\cdots,\boldsymbol{\mathcal{Z}}^{(m+n)}_{(\cdots:j_n)}).
\label{eq6}
\end{split}\end{equation}
The multi-indices rule obey little-endian convention, e.g., $\overline{i_1i_2\cdots i_N}=1+\sum_{k=1}^N (i_k-1)\prod_{m=1}^{k-1}I_m$. Note that the TN-ranks of $\boldsymbol{\mathcal{W}}$ can be obtained via (\ref{eq2}) in a similar manner. After that, the dense matrix $\boldsymbol{W}$ is represented as a multilinear operation with a few factors while retaining sufficient flexibility to carry out signal transformations. 

% Then the reduction ratio of in parameter ($P_{conv}$) and FLOPs ($C_{conv}$) are
% \begin{equation}
% \begin{split}
% \resizebox{0.9\hsize}{!}{$\begin{aligned}
% P_{conv}=\frac{K^2ST}{(2K+S+T)R^3}\quad  C_{conv}=\frac{K^2STWH}{ WH(S+T)R^3 + WHK^2R^5 + K^2R^5 }.
% \label{conv_conpl}
% \end{aligned}$}
% \end{split}
% \end{equation}

\vspace{2mm}
\noindent\textbf{ALS-based Scalable Compression.}
In fact, under a pre-specified target model size, the hyperparameters $\kappa$ and decomposed TN-ranks can be easily obtained by a binary search strategy~\cite{ref39} and Eq.(\ref{eq2}). This enables us to select well-suited TN-ranks separately for each instance weight tensor with constrained storage capacity. Subsequently, our goal is to obtain the weights of the decomposed network by minimizing the following optimization problem
\begin{equation}\begin{split}
\min_{\{\boldsymbol{\mathcal{Z}}^{(i)}_l \}_{i=1,l=1}^{N,L}}\ \ \sum_{l=1}^L \frac{1}{2} \Bigl|\Bigl| \boldsymbol{\mathcal{W}}_l-\Re(\boldsymbol{\mathcal{Z}}_l^{(1)},\cdots, \boldsymbol{\mathcal{Z}}_l^{(N)})   \Bigl|\Bigl|_F^2.
\label{eq3}
\end{split}\end{equation}
As described in Algorithm~\ref{alg1}, an ALS-based scalable compression algorithm is proposed to yield weights of a decomposed model in TN-format.

Overall, the STN comprises two separate stages (see Fig.~\ref{img2}), training and compression, without the concerns of fine-tuning and model selection and can be coupled with arbitrary architectures. The effectiveness of STN is fully verified in the next section.

% After decomposing a large pre-trained DNN model, we retrain the compressed network using the standard stochastic gradient descent (SGD) algorithm to obtain a smaller high-precision model in TN-format.

\begin{figure*}[!h]
  \centering
  \begin{minipage}[t]{0.32\textwidth}
  \includegraphics[width=1\linewidth]{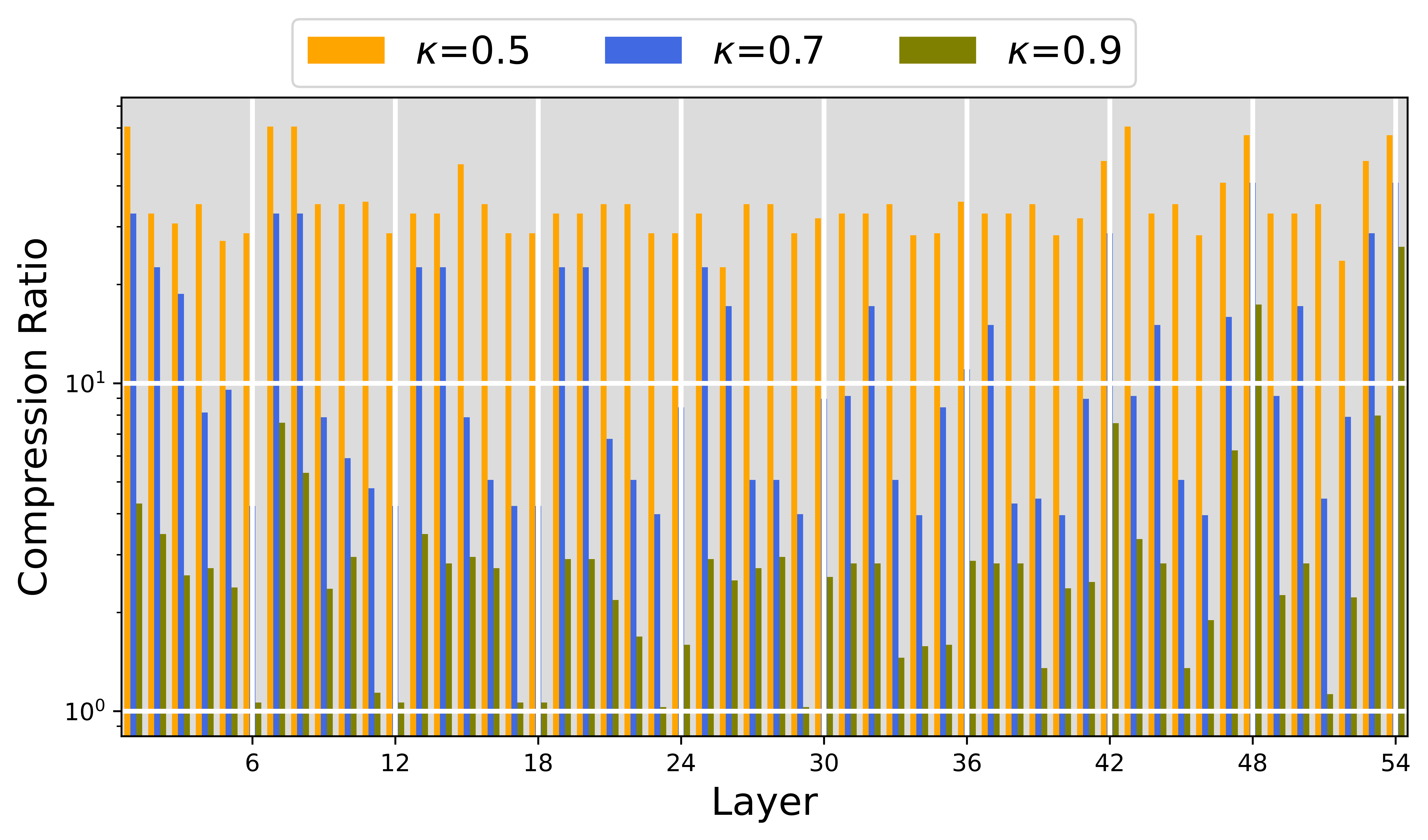}
  \footnotesize \subcaption{ Layer-wise Compression}
  \end{minipage}
  \begin{minipage}[t]{0.3\textwidth}
  \includegraphics[width=1\linewidth]{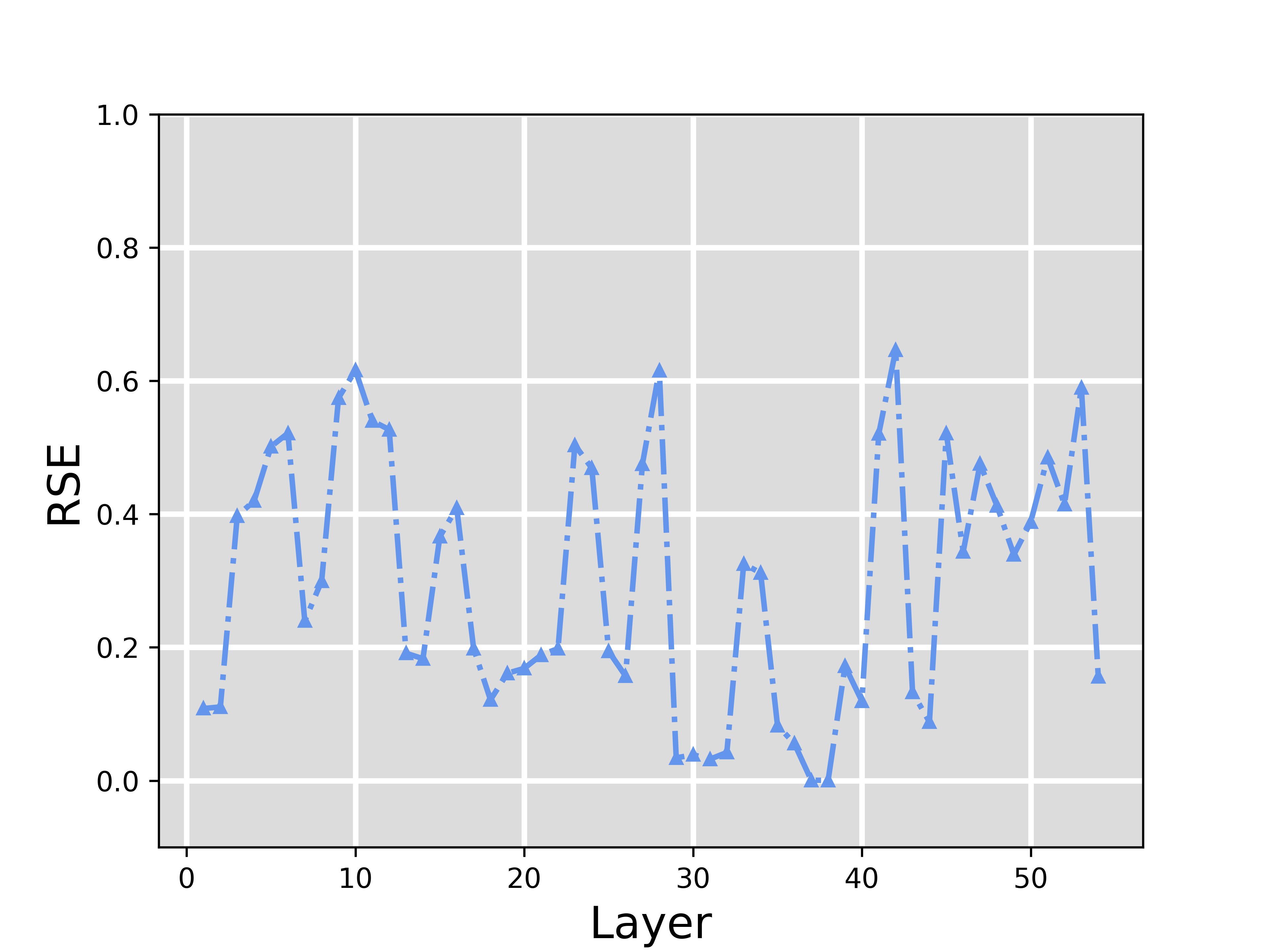}
  \footnotesize\subcaption{Approximation error}
  \end{minipage}
  \begin{minipage}[t]{0.3\textwidth}
  \includegraphics[width=1\linewidth]{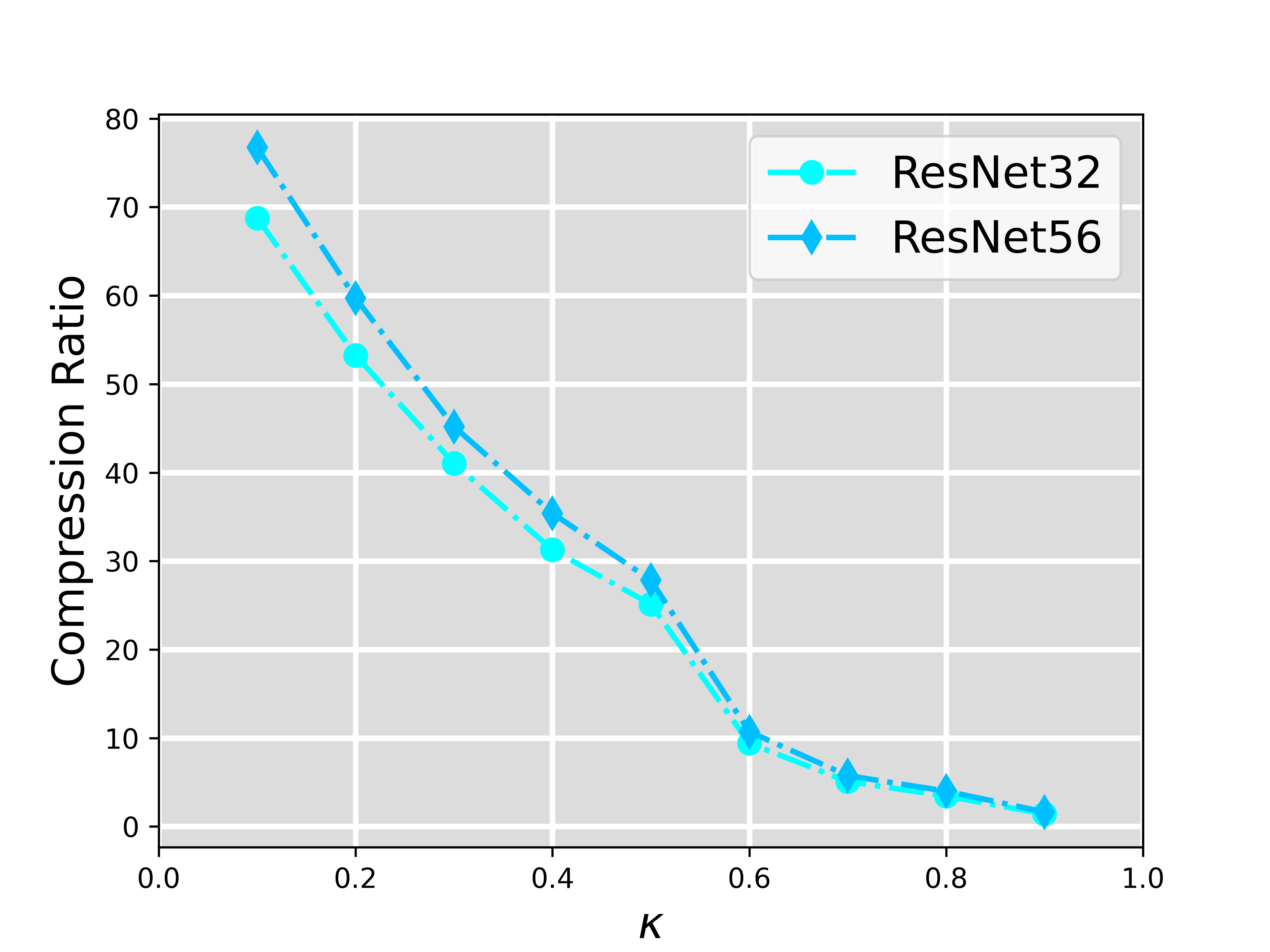}
  \centering
  \footnotesize\subcaption{Compression Ratio}
  \end{minipage}
  \begin{minipage}[t]{0.32\textwidth}
  \includegraphics[width=1\linewidth]{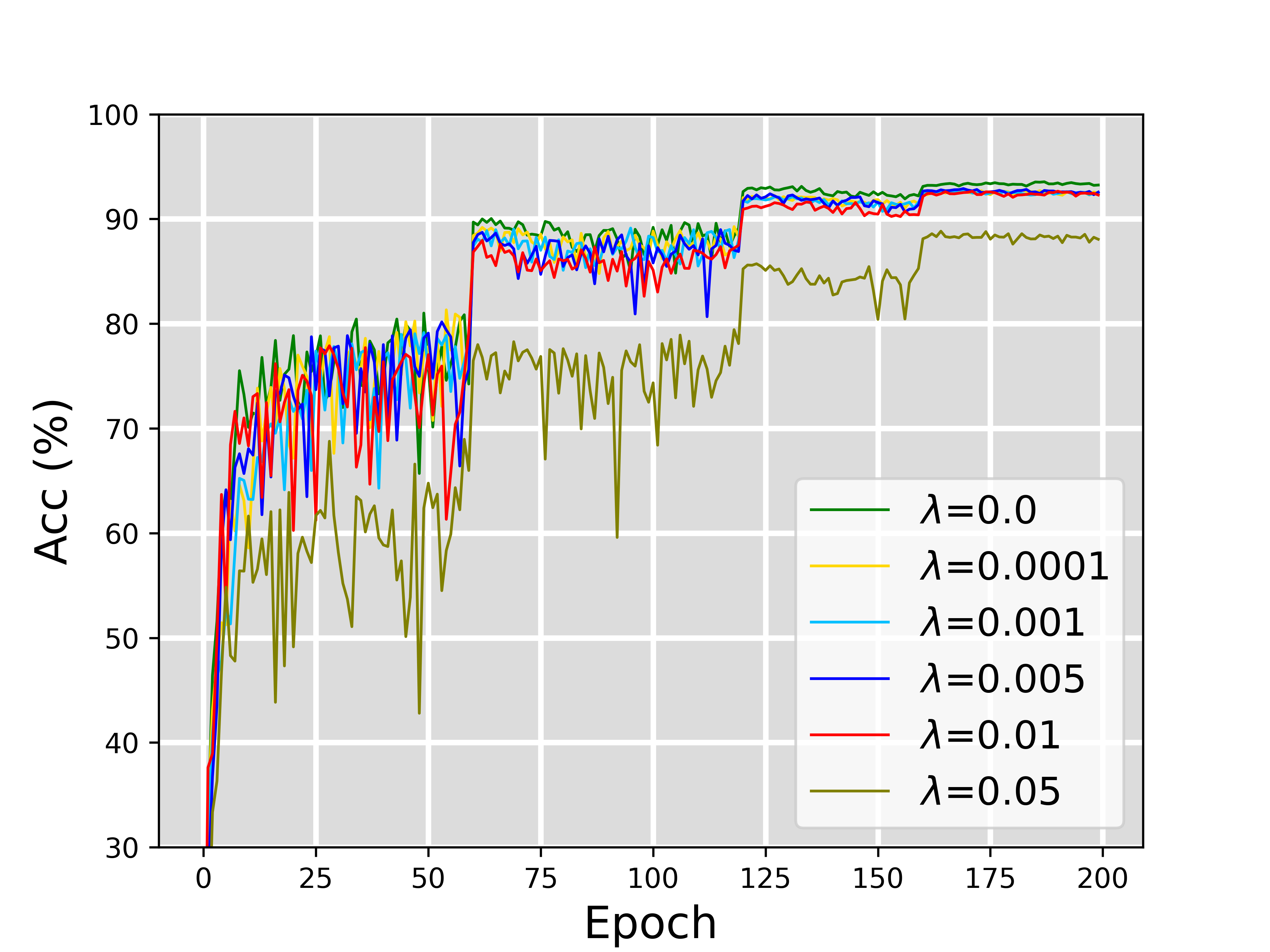}
  \centering
  \footnotesize\subcaption{Training Acc.}
  \end{minipage}
  \begin{minipage}[t]{0.32\textwidth}
  \includegraphics[width=1\linewidth]{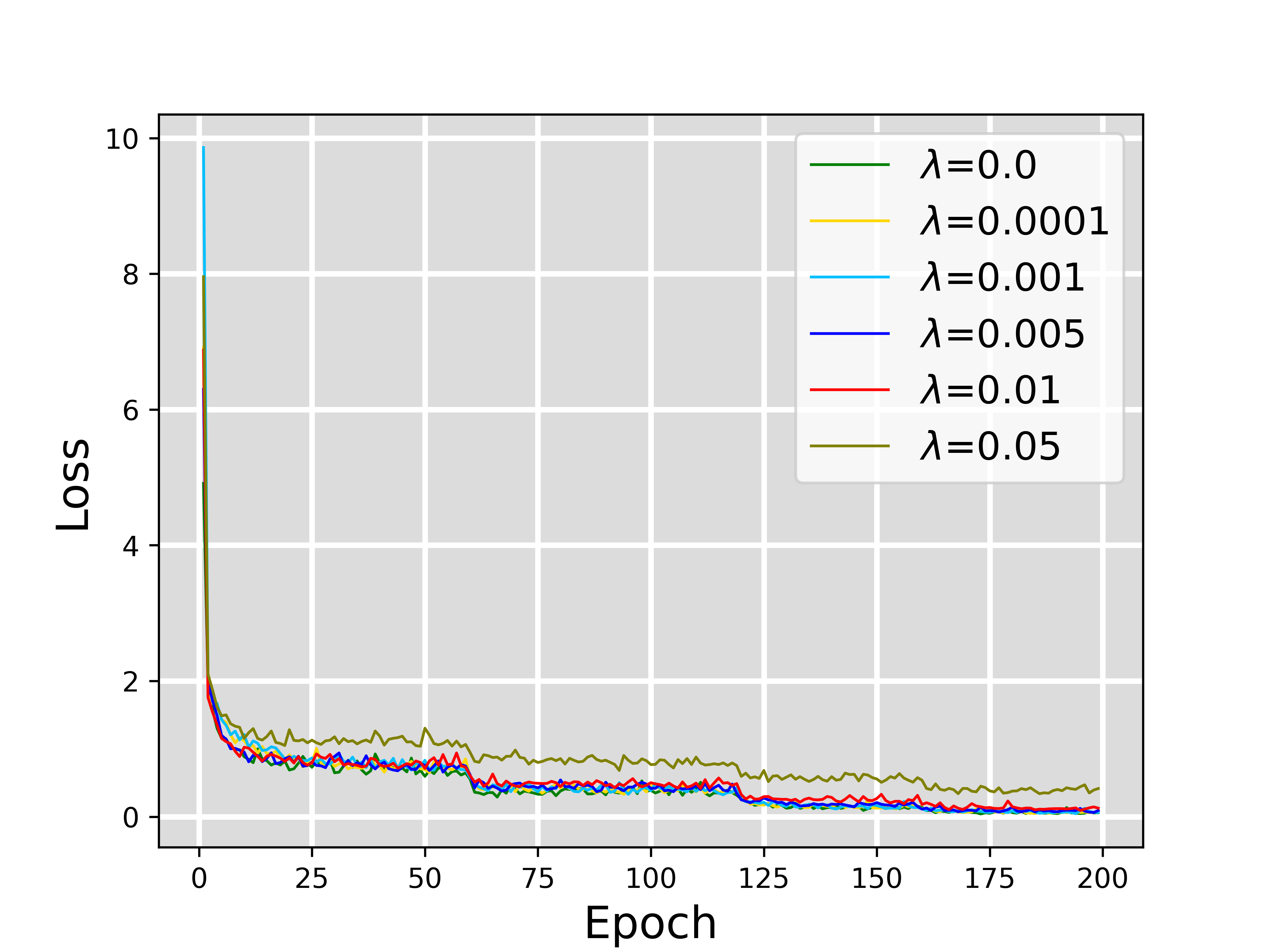}
  \centering
  \footnotesize\subcaption{Training Loss}
  \end{minipage}
  \begin{minipage}[t]{0.32\textwidth}
  \includegraphics[width=1\linewidth]{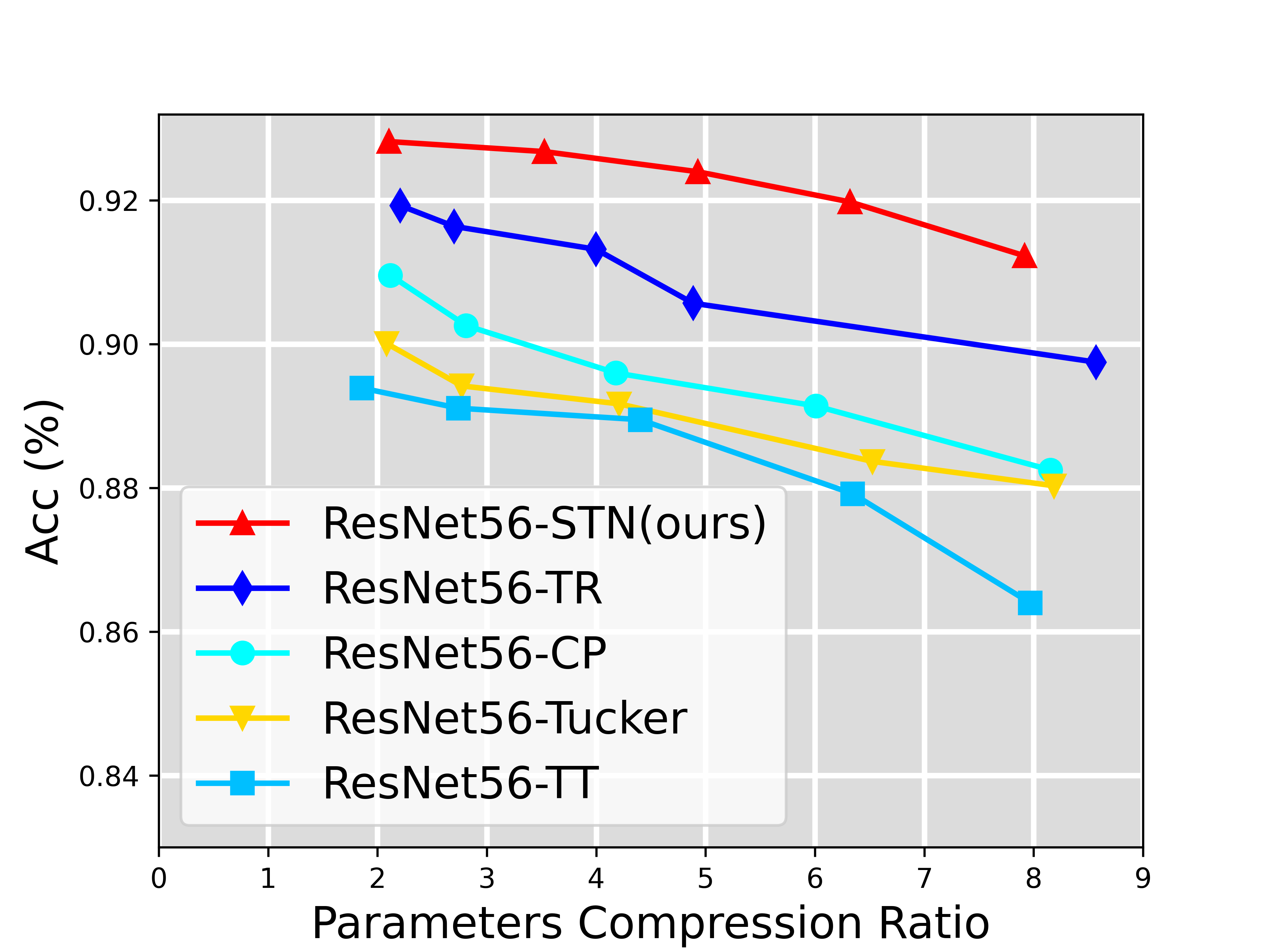}
  \centering
  \footnotesize\subcaption{Compression vs. Acc.}
  \end{minipage}
  \caption{Ablation study with ResNet-56 on CIFAR10. (a) Compression ratios of 54 convolutional layers under different $\kappa$ in decomposition stage. (b) Approximate relative square error (RSE) of convolutional layers of ResNet-56-STN with 4$\times$ compression. (c) Trade-off curve between model redunction ratios and $\kappa$. (d), (e) Test accuracy and training loss during structure-aware training stage with different $\lambda$. (f) Trade-off curves of accuracy vs. parameters compression ratio for TD-based ResNet-56 model. \vspace{-5mm}}
  \label{img3}
\end{figure*}

\vspace{-2mm}
\section{Experiments}
To validate the effectiveness of our methods, we conduct compression experiments on several popular architectures with different tasks. We apply our compression model on the ResNet family~\cite{ref49} for image classification tasks and report its validation accuracy on CIFAR10~\cite{ref50} and ImageNet datasets~\cite{ref51}. For the image segmentation task, we perform compression on the renowned U-Net architecture~\cite{ref52} and evaluate the performance on the medical Lung-CXR dataset~\cite{ref53}. In comparison, we evaluated other TD and quantization strategies-based networks on the same baselines, and other scalable approaches. For fairness, our experiments follow the same setups as~\cite{ref20,ref54}.

\vspace{-2mm}
\subsection{Ablation Study}
The tuning parameter $\kappa$ in (\ref{eq2}) controls the computational complexity and performance of STN during its compression stage. Note that we only use $\kappa$ to manipulate the size of the target model instead of manually designing the decomposition rank for each layer separately, which greatly simplifies the resizing of the model. After that, we experimentally reveal that different layers of a DNN model can exhibit various low-rank behaviours. As shown in Fig.~\ref{img3}(a), it displays the compression ratio of 54 convolutional layers inside BasicBlock of ResNet-56 under $\kappa$ of 0.5, 0.7, and 0.9, respectively. We can observe that different layers present significant compression differences even at the same $\kappa$, e.g., the 26th and 41st layers achieve 22$\times$ and 60$\times$ compression, respectivel, at $\kappa = 0.5$. Further, we visualize the RSE per convolutional layers of Resnet-56 in Fig.~\ref{img3}(b), where the compression ratios per layer are similar. It can be observed that the degree of approximation varies significantly among layers, which further provides empirical evidence that different layers exhibit apparent low-rankness behaviour inconsistencies across networks. Therefore, it is appropriate to select adaptive TN-ranks for each layer compression based on the intrinsic properties of the weight tensors, which enables us to reconfigure the model more flexibly and efficiently.

Next, we analyze the influence of the parameter $\lambda$ in (\ref{eq9}) on the model performance and convergence during the structure-aware training process of STN. As shown in Fig.~\ref{img3}(d), which illustrates the validation accuracy curves under different $\lambda$ of ResNet-56. We notice that the regularized model exhibits similar performance when $\lambda\leq 0.01$. In addition, the setting of $\lambda$ barely affects the training convergence of the regularized ResNet-56 (see Fig.~\ref{img3}(e)). The accuracy of the network, however, decreases rapidly when $\lambda=0.05$, attributed to the fact that the excessively low-rank constraint bounds the model's fitting capability. In order to obtain a simultaneous low-rank and high-accuracy model, we choose $\lambda=0.005$ in the subsequent experiments.

\begin{table}[!t]
\begin{minipage}[t]{0.48\textwidth}
\label{tab2}
  \centering
     \makeatletter\def\@captype{table}\makeatother\caption{Performance comparison of various methods based on TD and pruning strategies in compressing ResNet-18 on ImageNet. $\downarrow $ denotes compression ratio.}
\resizebox{1\textwidth}{!}{
        \begin{tabular}{l|c|c|c}
  \toprule
  \textbf{Method}& \textbf{Strategy} &\textbf{FLOPs $\downarrow $}  & \textbf{Top-5 (\%)}\\ 
  \hline
  \midrule
ResNet-18(baseline)~\cite{ref49} & -    & 1.0 $\times$   &   88.94    \\
\hline
DACP-ResNet-18~\cite{ref57}  & \multirow{4}{*}{Pruning}     & 1.9 $\times$   &   87.60    \\
FBS-ResNet-18~\cite{ref58}  &    & 2.0 $\times$   &   88.22    \\
FPGM-ResNet-18~\cite{ref56}  &   & 1.7 $\times$   &   88.53    \\
DSA-ResNet-18~\cite{ref59}  &    & 1.7 $\times$   &   88.35    \\
STN-ResNet-18 (ours) && \textbf{2.3 $\times$}   &    \textbf{88.65}  \\
\hline
SVD-ResNet-18~\cite{ref37} & \multirow{4}{*}{TD}     & 3.2 $\times$   &   86.61    \\
TT-ResNet-18~\cite{ref30} && \textbf{4.6 $\times$}   &   85.64    \\
TR-ResNet-18~\cite{ref31} && 4.3 $\times$   &   86.29    \\
STN-ResNet-18 (ours) && 4.5 $\times$   &    \textbf{87.86}  \\
    \bottomrule
 \end{tabular}}
\end{minipage}
  \begin{minipage}[t]{0.5\textwidth}
  \label{tab3}
   \centering
        \makeatletter\def\@captype{table}\makeatother\caption{Performance comparison in compressing U-Net on Lung-CXR. The evaluation indexes include the compression ratio of parameters (CR) and FLOPs, respectively, as well as the area under the precision-recall curve (PRAUC). }
         \resizebox{1\textwidth}{!}{
  \begin{tabular}{l|c|c|c}
  \toprule
  \textbf{Method} &  \textbf{CR} &\textbf{FLOPs $\downarrow $}  & \textbf{PRAUC(\%)}\\ 
  \hline
  \midrule
  U-Net(baseline)~\cite{ref52}   &1.0$\times$     &1.0$\times$   & 98.2\\
\hline
  DSC-U-Net~\cite{ref55}   &\textbf{7.9$\times$}     &5.2$\times$   & 97.1\\
  CP-U-Net~\cite{ref22}   &6.1$\times$     &4.2$\times$   & 97.4\\
  Tucker-U-Net~\cite{ref28}   &6.9$\times$     &4.6$\times$   & 96.4\\
  TT-U-Net~\cite{ref30}   &5.8$\times$     &4.0$\times$  & 97.5\\
  TR-U-Net~\cite{ref31}   &7.0$\times$     &5.1$\times$   & 97.8\\
  STN-U-Net (ours)  &6.5$\times$     &\textbf{5.5$\times$ }   & \textbf{98.0}\\
    \bottomrule
  \end{tabular}}
   \end{minipage}
\end{table}

\subsection{Comparison with Compression Methods}
In this part, we compare STN with other one-shot compression methods and conduct experiments on several benchmarks. For the fair competition, all tensorized networks are trained from scratch to verify the superiority of adaptive TN decomposition.

\noindent\textbf{CIFAR10.} We compare STN with other TD methods, including Tucker, CP, TT, and TR decomposition~\cite{ref22,ref28,ref30,ref31}. Figure \ref{img3}(f) plots the trade-off curves of accuracy versus parameters compression ratio for TD-based ResNet-56. Note here we only decompose the convolutional layers since the number of parameters in fully-connected layers is negligible. ResNet-56-STN can achieve higher accuracy with less storage cost, whereas other TD methods lead to significant performance degradation. For instance, even for the advanced TT and TR decompositions, they cause 3.9\% and 1.7\% accuracy loss of ResNet-56 at 4.4$\times$ and 4.0$\times$ compression, respectively. Surprisingly, our method achieves 4.1$\times$ and 3.5$\times$ parameters reduction on compressing ResNet-56 with only 0.6\% and 0.3\% performance degradation. This indicates that our adaptive approach achieves higher compression ratios and better performance than other TD methods.

\noindent\textbf{ImageNet.} Next, we conduct experiments on the large benchmark dataset ImageNet and report the experimental results of various TD- and pruning strategy-based methods towards compressing ResNet-18 in Table 1. Here we decompose the convolutional and fully-connected layers of ResNet-18 under $\kappa=0.8$, and obtain 4.5$\times$ FLOPs reduction at the expense of 1.08\% accuracy loss. Consequently, STN can achieve higher accuracy with similar parameters than TD-based methods and construct compacter parameter spaces compared with pruning-based methods. This experiment further demonstrates the superiority of STN, i.e., exploring the low-rank structure of each layer of the weight tensor to maximize compression.

\noindent\textbf{Lung-CXR.} We validate the generalization ability of STN on the image segmentation task. As shown in Table 2, we employ various TD methods and depthwise separable convolution (DSC)~\cite{ref55} to compress the prominent U-Net architecture~\cite{ref52}. The STN achieves 6.5$\times$ compression and best segmentation performance with the cost of 0.2\% drop of PRAUC. On the other hand, we avoid the hassle of setting the TN-ranks for each layer. Therefore, STN is more suitable for resizing the model flexibly and achieving better performance.

\subsection{Comparison with Scalable Networks}
We have demonstrated the superiority of adaptive decomposition in the previous experiments. Here, we further show the effectiveness of structure-aware training of STN. As shown in Fig.4, direct compressing the routinely trained full-rank ResNet-32 via Algorithm~\ref{alg1} leads to rapidly decreasing accuracy. In contrast, STN performing approximation on the low-rank model can maintain model performances well in a wide range. In particular, the model accuracy drops by only 0.9\% at a compression ratio of 2.1$\times$ (Table 3). It implies that introducing an additional low-rank regularizer during the training procedure can guarantee the desired low-rank characteristics of the network, thus effectively reducing the approximation error induced in the post-processing stage. More strikingly, the SVD-based Tucker decomposition with the variational Bayesian matrix factorization (VBMF) rank determination strategy performs even worse than the routinely trained STN. That hints iteratively decomposing the convolution kernel with the ALS algorithm has a lower RSE. Finally, the results in Table 4 further evidence the advantage of STN in inference speed.

\makeatletter\def\@captype{figure}\makeatother
\begin{minipage}{.45\textwidth}
\centering
  \includegraphics[width=1\linewidth]{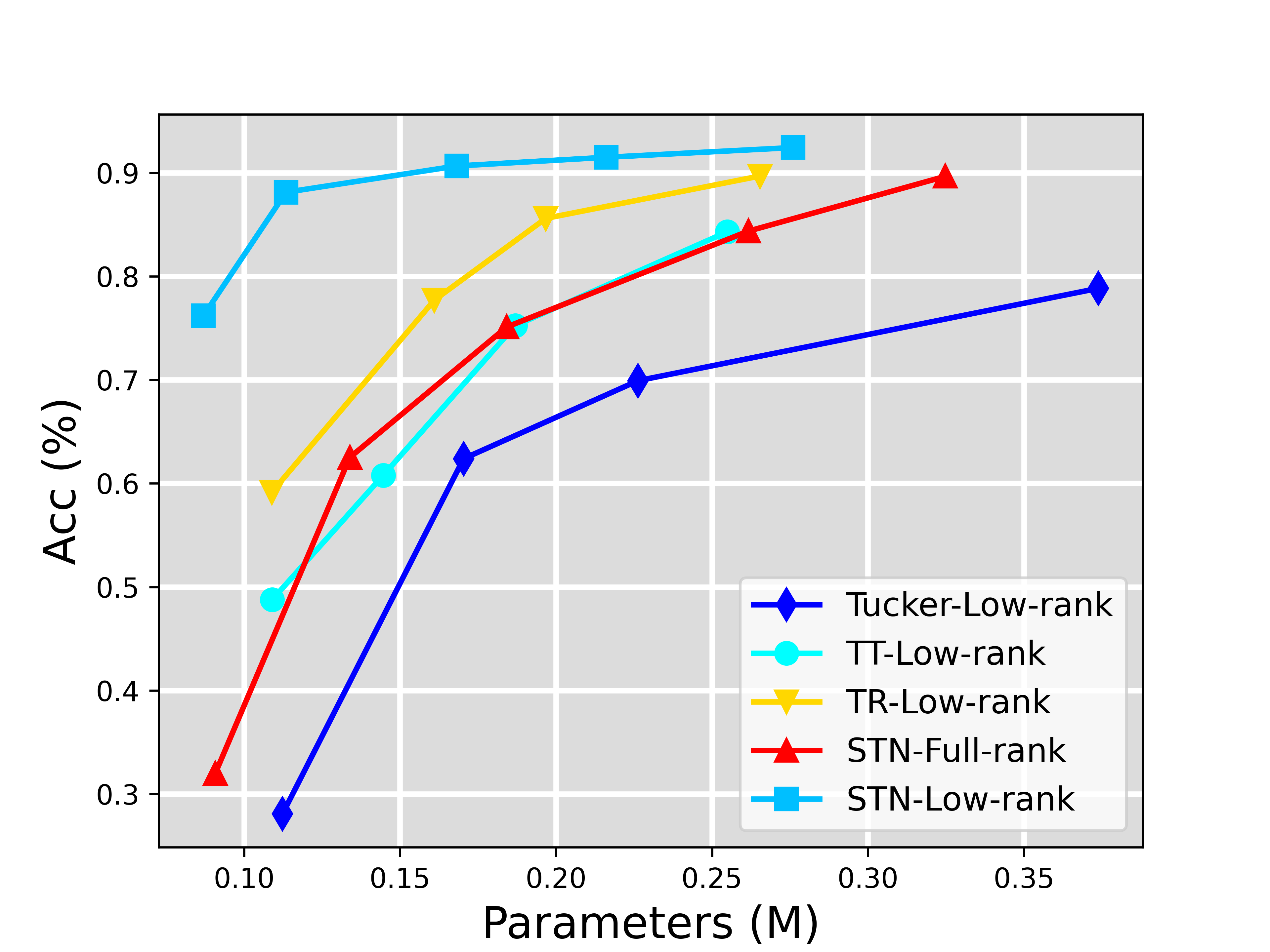}
  \caption{Scalable compression results with ResNet-32 on the CIFAR10 datasets.}
\end{minipage}
\makeatletter\def\@captype{table}\makeatother
\begin{minipage}{.45\textwidth}
   \centering
    \caption{Performance comparison of various TD methods in compressing ResNet-32 without fine-tuning.}
      \resizebox{1.0\textwidth}{!}{
  \begin{tabular}{l|c|c|c}
  \toprule
  \textbf{Method} &  \textbf{CR} &\textbf{FLOPs $\downarrow $}  & \textbf{Top-1 (\%)}\\ 
  \hline
  \midrule
  ResNet-32(baseline)~\cite{ref49}   &1.0$\times$     &1.0$\times$   & 92.5\\
\hline
  Tucker-Low-rank~\cite{ref28}   &{1.3$\times$}     &1.5$\times$   & 78.8\\
  TT-Low-rank\cite{ref30}   &1.8$\times$     &1.8$\times$  & 84.3\\
  TR-Low-rank~\cite{ref31}   &1.7$\times$     &2.0$\times$   & 89.7\\
  STN-Full-rank (ours)  &1.8$\times$     &{1.9$\times$ }   & {84.4}\\
  STN-Low-rank (ours)  &\textbf{2.1}$\times$     &\textbf{2.5$\times$ }   & \textbf{91.6}\\
    \bottomrule
  \end{tabular}}
\end{minipage}

\begin{figure*}[!t]
  \centering
   \includegraphics[width=0.7\linewidth]{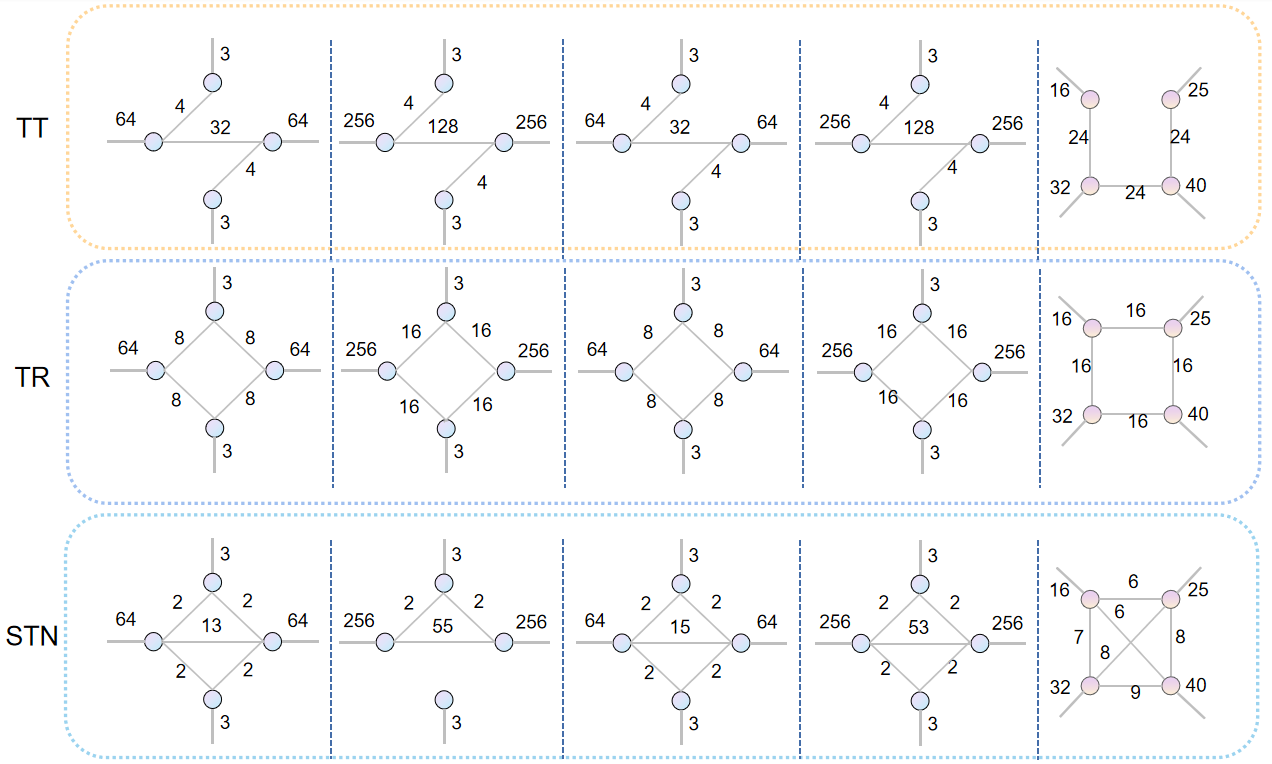}
   \caption{ Visualization of topologies and TN-ranks for compressing the 2nd, 6th, 10th, 14th, and 18th layers of ResNet-18 via TT, TR, and STN ($\kappa=0.7$). Note that TT and TR have fixed structures and manually determined ranks. The front four graphics correspond to convolutional layers, and the last one refers to fully-connected layers.\vspace{-5mm}}
   \label{img5}
\end{figure*}

\begin{table}[!t]
\begin{minipage}[t]{0.48\textwidth}
\centering
  \centering
     \makeatletter\def\@captype{table}\makeatother\caption{Comparison of inference speed (Intel-Xeon 6226R CPU) per image with ResNet-56 on the CIFAR10 datasets.}
\resizebox{1\textwidth}{!}{
  \begin{tabular}{l|c|c|c}
  \toprule
  \textbf{Method}& \textbf{Top-1 (\%)}      & \textbf{Time (ms)}       &\textbf{Speed}  \\ 
  \hline
  \midrule
ResNet-56 (baseline)~\cite{ref49} & 92.5  &4.22 &1.00 $\times$\\
\hline
CP-ResNet-56~\cite{ref22} & 90.26 &  2.86 & 1.47 $\times$\\
Tucker-ResNet-56~\cite{ref22} & 90.03& 2.06  & 2.05 $\times$\\
TT-ResNet-56~\cite{ref22} & 89.39 & 3.47 & 1.22 $\times$\\
TR-ResNet-56~\cite{ref22} & 90.57 & 2.90 & 1.45 $\times$\\
STN-ResNet-56 (ours) & \textbf{91.23}  & \textbf{1.83} & \textbf{2.31} $\times$\\
    \bottomrule
  \end{tabular}}
   \end{minipage}
\begin{minipage}[t]{0.48\textwidth}
\label{tab4}
  \centering
     \makeatletter\def\@captype{table}\makeatother\caption{Performance comparison of various scalable methods in compressing ResNet-18 on ImageNet.}
\resizebox{1\textwidth}{!}{
  \begin{tabular}{l|c|c|c}
  \toprule
  \textbf{Method}& \textbf{Top-1 (\%)}      & \textbf{Top-5 (\%)}       &\textbf{CR}  \\ 
  \hline
  \midrule
ResNet-18(baseline)~\cite{ref49} & 69.10    & 88.94 $\times$   &   1.0 $\times$    \\
\hline
TRP1+Nu~\cite{ref37}   &  65.39 & 86.37 & 2.2 $\times$\\
 $[$Jaderberg $et\ al$.$]$ \cite{ref11}  &  62.80 & 83.72 & 2.0 $\times$\\
$[$Zhang $et\ al$.$]$ \cite{ref60}  &  63.10 & 84.44 & 1.4 $\times$\\
Slimmable-net~\cite{ref47}    &  66.52 & 87.12 & 1.9 $\times$\\
Decomposable-net~\cite{ref27}    &  67.14 & 87.30 & 2.3 $\times$\\
STN (ours) &\textbf{ 67.85}  &\textbf{ 88.03}  &    2.2 $\times$  \\
    \bottomrule
  \end{tabular}}
  \end{minipage}
 \vspace{-5mm}
\end{table}

We compare STN with other scalable methods, including TRP~\cite{ref37} , Slimmable networks~\cite{ref47}, and Decomposable-net~\cite{ref27}. Table 5 reports the performance of various approaches towards compressing ResNet-18 on ImageNet. Obviously, STN achieves the best 67.85\% classification accuracy with 2.2$\times$ compression. Compared to TRP, the precision is 2.46 higher at the same compression ratio. This implies that the STN trained under the ADMM framework possesses low-rank characteristics and facilitates the subsequent adaptive decomposition. More interestingly, we visualize the decomposed topological structures and TN-ranks of several convolutional and the fully-connected layers of ResNet-18 in Fig.~\ref{img5}, and it is evident that the adaptiveness of STN includes two parts: decomposition mode and rank selection. Also, we can see the intensity of the mode correlation of the weight tensors according to the magnitude of the edge rank between the connected factors; for example, the correlation between the channel modes of convolution kernels is stronger than the spatial ones. Moreover, the topological structure easily forms a fully-connected graph for the fully-connected layer due to its balanced mode dimension. In conclusion, the adaptive compression of the weight parameters and the low-rank structure-aware training is crucial for the success of the STN.

\vspace{-5mm}
\section{Conclusion}
We propound the Scalable Tensorizing Networks (STN) to improve parameter efficiency and scalability of neural networks. First, we introduce a low-rank regularizer during the training of the DNN model and develop a structure-aware training algorithm, which can effectively reduce the approximation error of the decomposition and thus avoid the subsequent retraining stage. After that, recognizing that different layers exhibit various low-rank behaviours and ALS-based adaptive compression strategy is adopted to compress the convolutional and fully-connected layers simultaneously. Comprehensive experiments on multiple network architectures substantiate the superiority and effectiveness of our approach over other state-of-the-art low-rank and scalable networks.

\section{Appendix}
\subsection{Analysis}

We report the storage and calculation complexity for different methods against compressed convolutional and fully-connected layers in Tables 1 and 2. Note that the \textbf{CP and Tucker decompositions are not suited for fully-connected layer compression}. Despite the higher theoretical complexity of our STN, empirical evidence in the main paper has revealed that adaptive compression is more advantageous in dealing with mode-unbalanced low-order tensors, and the compressed DNNs achieve higher performance. In addition, the split dimension in the tensorization process obeys the variance minimization principle.

\begin{table*}[ht]
\begin{center}
\caption{Comparison of parameters and FLOPs involved in convolutional layer compressed by different methods.}
\label{tab1}
\resizebox{0.8\textwidth}{!}{
\begin{tabular}{c|c|c}
\hline
Methods   & Parameters  &  FLOPs\\
\hline
Conv (Standard) & $K^2\times S \times T$  & $K^2\times S\times T\times W\times H$ \\
DSC~\cite{ref55} & $(K^2+T)\times S$   & $(K^2+T)\times S  \times W\times H$ \\

CP~\cite{ref22}    & $R_{cp}(S+T+2K)$     & $R_{cp}(C+D+2K)\times W\times H$  \\
Tucker~\cite{ref28}    & $R_{Tuc}(S+T) + R_{Tuc}^2K^2 $      & $ R_{Tuc}(C+D+R_{Tuc}K^2) \times W\times H $      \\

TT~\cite{ref30}    & $ (S+T +  2K R_{TT}) R_{TT}  $    & $(S+T + K^2R_{TT})R_{TT}\times W\times H + K^2R_{TT}^3   $  \\
TR~\cite{ref31}    & $ (S+T+2K) R_{TR}^2  $      & $ (S+T + K^2R_{TR}^2)R_{TR}^2\times W\times H + K^2R_{TR}^4   $      \\

STN (ours)        & $(S+T+2K) R_{TN}^3$      & $ (S+T + K^2R_{TN}^2)R_{TN}^3\times W\times H + K^2R_{TN}^5 $      \\
\hline
\end{tabular}}
\end{center}
\end{table*}

\begin{table*}[h]
\begin{center}
\caption{Comparison of parameters and FLOPs involved in fully-connected layer compressed by different methods. The weight matrix $\boldsymbol{W}\in \mathbb{R}^{M\times N}$ and $M=I_1I_2, N=J_1J_2$.}
\label{tab2}
\begin{tabular}{c|c|c}
\hline
Methods   & Parameters  &  FLOPs\\
\hline
FC (Standard) & $I_1I_2\times J_1J_2$  & $I_1I_2\times J_1J_2\times C$ \\

TT~\cite{ref30}    & $(I_1 + J_2) R_{TT} + (I_2 + J_1) R_{TT}^2$   &   $(I_1I_2+ J_1J_2) R_{TT} \times (C+R_{TT}) $ \\
TR~\cite{ref31}    & $(I_1+ I_2+J_1+ J_2) R_{TR}^2$    &    $(I_1I_2+ J_1J_2) R_{TT}^2 \times (C+R_{TT}) $  \\

STN (ours)        &  $(I_1+ I_2+J_1+ J_2) R_{TN}^3$    &    $(I_1I_2+ J_1J_2) R_{TN}^3\times (C+R_{TN}^2) $  \\
\hline
\end{tabular}
\end{center}
\end{table*}

The storage and computation complexity of TD-based compression is closely related to the rank selection, and we hereby reveal the relationship between the TN-rank with the CP and Tucker ranks in Theorem 1.

\vspace{2mm}
\noindent\textbf{Theorem 1.}{\it
\ \ Consider an $N$th-order tensor $\boldsymbol{\mathcal{X}}\in\mathbb{R}^{I_1\times\dots\times I_N}$ can be represented in TN-format. Then we have: (1) If $\boldsymbol{\mathcal{X}}$ has CP rank $R_{cp}$, then $rank(\boldsymbol{X}_{[n_{1:r};n_{r+1:N}]})\leq \min\{\prod_{i=1}^r \prod_{j=r+1}^N R_{n_i,n_j}, R_{cp}\} $. (2) If $\boldsymbol{\mathcal{X}}$ has Tucker rank $R_{Tuc}=(r_1,\cdots,r_N)$, then $rank(\boldsymbol{X}_{[n_{1:r};n_{r+1:d}]})\leq \min\{\prod_{i=1}^r r_{n_i}, \prod_{j=r+1}^N r_{n_j} \}$,}, where $\boldsymbol{X}_{[n_{1:r};n_{r+1:d}]} \in \mathbb{R}^{I_{n_1}\dots I_{n_r}\times I_{n_{r+1}}\dots I_{n_N}}$.

\subsection{More experimental results}
We illustrate the trade-off curve of STN towards compressing ResNet-50 on the CIFAR100 datasets in Figure 1. Compared with the TR-based compression (TR-ResNet-50) and trained rank pruning (STN-ResNet-50) method~\cite{ref37}, the experimental results further validate the superiority of adaptive compression and scalable strategies.

\begin{figure}[!ht]
  \centering
   \includegraphics[width=1\linewidth]{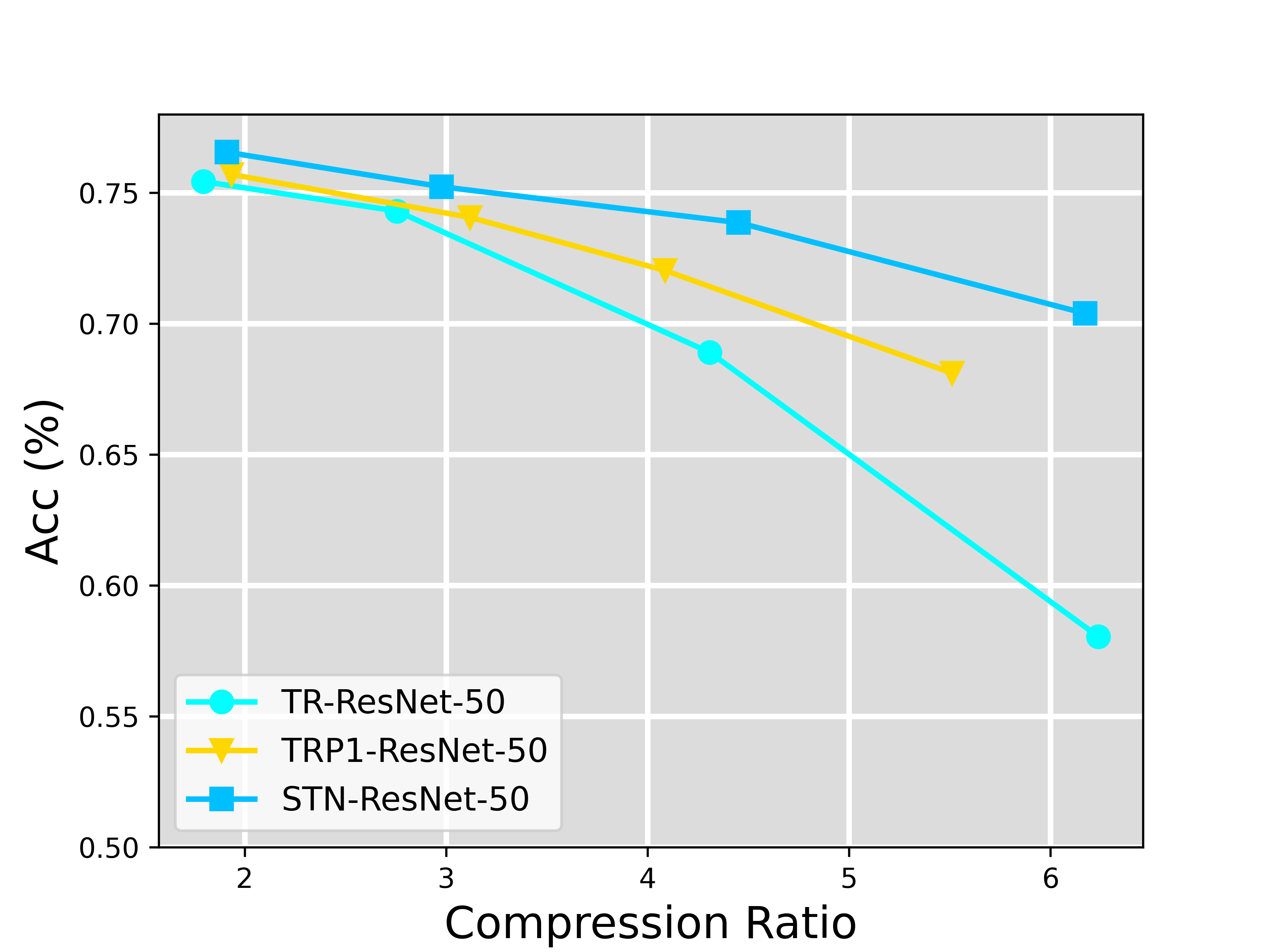}
   \caption{Scalable compression results with ResNet-50 on the CIFAR100 datasets. Our STN-ResNet50 consistently achieves the best performance at varying parameter compression ratios}
   \label{img5}
\end{figure}

%%%%%%%%% REFERENCES
{\small
\bibliographystyle{ieee_fullname}
\bibliography{egbib}
}

\end{document}